\documentclass{article}

\usepackage{arxiv}

\usepackage[utf8]{inputenc} 
\usepackage[T1]{fontenc}    
\usepackage{hyperref}       
\usepackage{url}            
\usepackage{booktabs}       
\usepackage{amsfonts}       
\usepackage{nicefrac}       
\usepackage{microtype}      
\usepackage{graphicx}
\usepackage{natbib}
\usepackage{doi}
\usepackage{color}
\usepackage{multirow}

\usepackage{mathtools}
\usepackage{amssymb}
\usepackage{latexsym}
\usepackage{dsfont}
\usepackage{mathrsfs}
\usepackage{amssymb}
\usepackage{amsfonts}
\usepackage{bm}
\usepackage{xspace}
\usepackage{amsthm}

\ifx\BlackBox\undefined
\newcommand{\BlackBox}{\rule{1.5ex}{1.5ex}}  
\fi

\ifx\QED\undefined
\def\QED{~\rule[-1pt]{5pt}{5pt}\par\medskip}
\fi

\ifx\proof\undefined
\newenvironment{proof}{\par\noindent{\em Proof:\ }}{\hfill\BlackBox\\[.0mm]}
\fi

\ifx\theorem\undefined
\newtheorem{theorem}{Theorem}[section]
\fi

\ifx\example\undefined
\newtheorem{example}{Example}[section]
\fi

\ifx\property\undefined
\newtheorem{property}{Property}
\fi

\ifx\lemma\undefined
\newtheorem{lemma}{Lemma}[section]
\fi

\ifx\proposition\undefined
\newtheorem{proposition}{Proposition}[section]
\fi

\ifx\remark\undefined
\newtheorem{remark}{Remark}
\fi

\ifx\corollary\undefined
\newtheorem{corollary}{Corollary}[section]
\fi

\ifx\definition\undefined
\newtheorem{definition}{Definition}[section]
\fi

\ifx\conjecture\undefined
\newtheorem{conjecture}{Conjecture}[section]
\fi

\ifx\axiom\undefined
\newtheorem{axiom}[theorem]{Axiom}
\fi

\ifx\claim\undefined
\newtheorem{claim}[theorem]{Claim}
\fi

\ifx\assumption\undefined
\newtheorem{assumption}{Assumption}
\fi

\ifx\problem\undefined
\newtheorem{problem}{Problem}[section]
\fi

\ifx\fact\undefined
\newtheorem{fact}{Fact}
\fi

\title{First Place Solution of \\ KDD Cup 2021 \& OGB Large-Scale Challenge \\ Graph Prediction Track}

\author{ 

Chengxuan Ying \\
Dalian University of Technology\\
\texttt{yingchengsyuan@gmail.com} \\
\And
Mingqi Yang \\
Dalian University of Technology\\
\texttt{yangmq@mail.dlut.edu.cn} \\
\And
Shuxin Zheng\thanks{Contact Person.} \\
Microsoft Research Asia\\
\texttt{ shuz@microsoft.com} \\
\And
Guolin Ke \\
Microsoft Research Asia\\
\texttt{guoke@microsoft.com } \\
\And
Shengjie Luo \\
Peking University\\
\texttt{luosj@stu.pku.edu.cn} \\
\And

Tianle Cai \\
Princeton University\\
\texttt{tianle.cai@princeton.edu } \\
\And
Chenglin Wu \\
Xiamen University\\
\texttt{ether.wcl@gmail.com } \\
\And
Yuxin Wang \\
Dalian University of Technology\\
\texttt{wyx@dlut.edu.cn } \\
\And
Yanming Shen \\
Dalian University of Technology\\
\texttt{shen@dlut.edu.cn } \\
\And
Di He \\
Microsoft Research Asia\\
\texttt{dihe@microsoft.com } \\

}

\renewcommand{\undertitle}{MachineLearning Team}

\hypersetup{
pdftitle={A template for the arxiv style},
pdfsubject={q-bio.NC, q-bio.QM},
pdfauthor={David S.~Hippocampus, Elias D.~Striatum},
pdfkeywords={First keyword, Second keyword, More},
}

\begin{document}
\maketitle

\begin{abstract}
	In this technical report, we present our solution of KDD Cup 2021 OGB Large-Scale Challenge - PCQM4M-LSC Track. We adopt Graphormer and ExpC as our basic models. We train each model by 8-fold cross-validation, and additionally train two Graphormer models on the union of training and validation sets with different random seeds. For final submission, we use a naive ensemble for these 18 models by taking average of their outputs. Using our method, our team MachineLearning achieved 0.1200 MAE on test set, which won the first place in KDD Cup graph prediction track.
\end{abstract}

\keywords{Graphormer \and ExpC \and Graph Neural Networks \and Graph Representation \and OGB-LSC}

\section{Datasets}

Using fast and accurate ML models to approximate Density Functional Theory (DFT) enables diverse downstream applications, such as property prediction for organic photovaltaic devices and structure-based virtual screening for drug discovery.

PCQM4M-LSC is a recent public dataset on the OGB Large-Scale Challenge (OGB-LSC)~\citep{hu2021ogb}, to encourage the development of state-of-the-art graph ML models. In particular, it is a quantum chemistry dataset aiming to predict DFT-calculated HOMO-LUMO energy gap of molecules given their 2D molecular graphs. Mean Absolute Error (MAE) is used as evaluation metric. 

There are about 3.8M graphs in PCQM4M-LSC, and we follow the official data split with ratio 80/10/10 using scaffold split. Each node is associated with a 9-dimensional feature (e.g., atomic number, chirality) and each edge comes with a 3-dimensional feature (e.g., bond type, bond stereochemistry). 

The union of training and validation sets is used for training. We do not use any external data for this challenge.

\section{Features}

We train ExpC using the default node and edge features. Except for the officially provided features, we additionally extract a number of domain-specific molecular features for training Graphormer, as shown in Table 1.

\begin{table}[ht]
    \centering
\begin{tabular}{l|ll}
\hline
Type                   & Attribute type            & Description                                          \\ \hline
\multirow{13}{*}{Atom} & Atomic number             & Number of protons                                    \\
                       & Degree                    & With Hydrogens and without Hydrogens                 \\
                       & Number of Hydrogens       &                                                      \\
                       & Hybridization             & Sp, sp2 or sp3 etc.                                  \\
                       & Aromatic atom             & a part of an aromatic ring                           \\
                       & Is in ring                &                                                      \\
                       & Valence                   & Explicit valence, implicit valence, total valence    \\
                       & Radical electrons         &                                                      \\
                       & Formal charge             &                                                      \\
                       & Gasteiger charge          &                                                      \\
                       & Periodic table features   & rvdw, default valence, outer electrons, rb0 and etc. \\
                       & Chirality                 & Is chiral center                                     \\
                       & Donor or accepter         & donate electron or accept electron                   \\ \hline
\multirow{5}{*}{Bond}  & Bond type                 & Single, double, triple, aromatic bond, etc.          \\
                       & Bond stereo               & Z, E, cis, trans double bond, etc.                   \\
                       & Bond direction            & Bond's direction (for chirality)                     \\
                       & Is conjugated             &                                                      \\
                       & Is in ring                &                                                      \\
                       & Euclidean distance        & Using MMFF optimizer (RDKit\footnote{\url{https://www.rdkit.org/docs/source/rdkit.Chem.html}}) to obtain the coordinates of a molecule        \\ \hline
Atom Pair              & Euclidean distance        & Using MMFF optimizer to obtain the coordinates of a molecule        \\ \hline

\end{tabular}

    \caption{Atomic and bond attributes used to construct graph inputs.}
    \label{tab:feat}
\end{table}

Table \ref{tab:feat} summarizes all extra features used in training Graphormer. We mainly reuse the features extracted by ~\citet{liu2021transferable} and ~\citet{helix2021}. Besides, to calculate the accurate HOMO-LUMO gap of a molecular graph, one should firstly use DFT to do geometric optimization for the corresponding molecular conformation, i.e., the 3D molecular structure. The geometric optimization calculated by DFT is more accurate but quite time-consuming, since it's based on quantum mechanics force field. An alternative choice is to use RDKit to calculate the 3D molecular structure (Euclidean distance in Table \ref{tab:feat}), which is based on classical mechanics force field, and is much faster but less accurate. Although the uncertainty is relatively large, we believe the RDKit-calculated molecular conformation could provide helpful information to help the model better predict the HOMO-LUMO gap. Therefore, we calculate the 3D conformation of each molecule by RDKit, and extract two features to represent the molecular structure: (1) the Euclidean distance of each bond as new edge feature; (2) the Euclidean distance of each atom pair as new node pair feature.

Feature extraction process has been parallelized to speed up. To extract all the features mentioned in Table \ref{tab:feat} for the entire PCQM4M dataset with 3.8$M$ molecules, it takes about 12 hours on the machine with single Intel(R) Xeon(R) CPU E5-2690 v4 @ 2.60GHz CPU. And the pre-processing over the 377$K$ test molecules can be done in less than 1.5 hour.

\section{Models}
In this section, we describe our two models used in this challenge.

\subsection{Graphormer}
We develop a powerful graph neural network called Graphormer for this challenge, which is elaborated in ~\cite{ying2021transformers}. We strictly follow the implementation described in the original paper. We  train a Graphormer with $L=12, d=784$, which has about 47M params.

Since the Euclidean distance calculated by RD-kit between atoms on the 3D molecular graph is more informative for characterizing the molecular properties than the shortest path distance, we choose the Euclidean distance as the $\phi(v_i,v_j)$ for the spatial encoding in Graphormer. Nevertheless, we found that the Euclidean distance feature does not generalize well on validation set, i.e., the model is more easily trapped into the over-fitting problem. One possible reason is that the model memorizes each molecule in training set through the Euclidean distances of all node pairs. To address this problem, we employ a Gaussian Radial Basis Function (RBF) kernel to kernelize the distance feature~\citep{unke2019physnet}. In our final solution, we use 256 kernels in RBF to improve the generalization ability of Graphormer.

Except for the spatial encoding, the Euclidean distance feature is also adopted in the edge feature to identify the length of bond (which reflects the strength of each bond, the shorter, the stronger). To further avoid the over-fitting problem, we modify the bond distance $X_{bonddist}$ by 

$$X_{bonddist}=X_{bonddist} + \text{Lapalce}(\mu,b),$$

where $\mu=0.001994, b=0.031939$. We choose $\mu$ and $b$ by fitting the difference between the calculated results of RDKit and DFT, on another dataset called QM9~\citep{ramakrishnan2014quantum}, which provides the DFT-calculated 3D molecular structures.

\subsection{ExpC$^{\star}$}
ExpC$^{\star}$ is a variant of our ExpandingConv framework in ~\cite{yang2020breaking}. We made some modifications to better fitting the given task.

\textbf{Neighborhood aggregation layer.} We use the following implementation to iteratively compute node representations layer by layer.
\begin{align}
    & \mathbf m^{(t)}_{uv} = \sigma(\mathbf W^{(t)}_1\mathbf h_{e_{uv}}) \nonumber\\
    & \mathbf h^{\prime(t-1)}_u = \sigma(\mathbf W^{(t)}_2\mathbf h_u^{(t-1)}),\ \mathbf h^{\prime(t-1)}_v = \sigma(\mathbf W^{(t)}_2\mathbf h_v^{(t-1)}) \nonumber\\
    & \mathbf h^{(t)}_{v} = \textrm{MLP}^{(t)}(\sum_{u\in\mathcal N(v)}\mathbf m^{(t)}_{uv}\odot\mathbf h^{\prime(t-1)}_u+\mathbf h^{\prime(t-1)}_v), \nonumber
\end{align}
where $\mathbf h^{(0)}_v=\mathbf x_v$ is the node $v$'s feature.
$\mathbf h_{e_{uv}}$ denotes edge feature among node pair $u$ and $v$.
$\odot$ denotes element-wise product.
$\mathbf W_1,\mathbf W_2\in\mathbb R^{d^{\prime}\times\star}$ are trainable matrices and $\textrm{MLP}:\mathbb R^{d^{\prime}}\rightarrow\mathbb R^d$ is a 2-layer perception where $d$ and $d^{\prime}$ are manually setting parameters.
We implement the aggregation coefficient computation $\mathbf m_{uv}$ introduced in ~\cite{yang2020breaking} as the function of edge features $\mathbf h_{e_{uv}}$ to leverage edge features in aggregations.

A neighborhood aggregation layer operates on sets of neighbors while preserving permutation invariance. Its distinguishing strength leads to the fundamental limitation of GNN performance as studied in ~\cite{zaheer2017deep,xu2018how,corso2020principal,yang2020breaking}.
In ~\cite{yang2020breaking}, the distinguishing strength analysis is restricted in linear space. Here, we simplify the problem by introducing a nonlinear process.
Specifically, by setting a larger $d^{\prime}$ ($d^{\prime}>d$ usually) for $\mathbf W_1$ and $\mathbf W_2$ and then applying element-wise nonlinear functions $\sigma$ before aggregations, we can achieve more powerful aggregations and also break the distinguishing strength bottleneck of SUM.
We will give a detailed investigation of this part in our future work.

\textbf{Readout.} 
The virtual node strategy \citep{hu2020open} can be considered as adding a fully connected node.
By introducing the virtual node, the aggregation on the virtual node can be used as a Readout implementation to generate the entire graph representation since it aggregates representation of all nodes.
We adopt the similar aggregation implementation as defined above on the virtual node.
Finally, we summarize the virtual node representations of all layers to generate the final graph representation to reduce the information loss of local structures in a deep model \citep{xu2018representation}.

\section{Training Strategies}

\paragraph{Graphormer.}
We report the detailed hyper-parameter settings used for training Graphormer in Table \ref{tab:gf_details}. The embedding dropout ratio is set to 0.1 by default in many previous Transformer works~\cite{devlin2019bert,liu2019roberta}. However, we empirically find that a small embedding dropout ratio (e.g., 0.1) would lead to an observable performance drop on validation set of PCQM4M-LSC. One possible reason is that the molecular graph is relative small (i.e., the median of \#atoms in each molecule is about 15), making graph property more sensitive to the embeddings of each node. Therefore, we set embedding dropout ratio to 0 on this dataset.

\begin{table*}[ht]
\centering 
\caption{Model Configurations and Hyper-parameters of Graphormer on PCQM4M-LSC. } \label{tab:gf_details}

\begin{tabular}{lcc}
\toprule
& Graphormer \\ \hline
\textbf{\#Layers} & 12  \\ 
\textbf{Hidden Dimension $d$} & 768  \\ 
\textbf{FFN Inner-layer Dimension} & 768  \\ 
\textbf{\#Attention Heads} & 32  \\ 
\textbf{Hidden Dimension of Each Head} & 24  \\ 
\textbf{FFN Dropout} & 0.1  \\ 
\textbf{Attention Dropout} & 0.1 \\ 
\textbf{Embedding Dropout} & 0.0 \\ 
\textbf{Max Steps} & 1.5$M$ \\
\textbf{Peak Learning Rate} & 2e-4 \\ 
\textbf{Batch Size} & 1024 \\ 
\textbf{Warm-up Steps} & 10$K$ \\ 
\textbf{Learning Rate Decay} & Linear  \\ 
\textbf{Adam $\epsilon$} & 1e-8 \\ 
\textbf{Adam ($\beta_1$, $\beta_2$)} & (0.9, 0.999) \\ 
\textbf{Gradient Clip Norm} & 5.0  \\ 
\textbf{Weight Decay} & 0.0  \\ 
\bottomrule
\end{tabular}

\end{table*}

\paragraph{ExpC$^{\star}$.}

\begin{table*}[ht]
\centering 
\caption{Model Configurations and Hyper-parameters of ExpC$^{\star}$ on PCQM4M-LSC. } \label{tab:expc_details}
\begin{tabular}{lcc}
\toprule
& ExpC$^{\star}$ \\ \hline
\textbf{\#Layers}                                   & 5  \\ 
\textbf{Hidden Dimension $d$}                       & 600  \\ 
\textbf{Expanding Hidden Dimension $d^{\prime}$}    & 1200 \\
\textbf{Dropout}                                    & 0.0  \\ 
\textbf{Max Epochs}                                 & 100 \\ 
\textbf{Batch Size}                                 & 256 \\ 
\textbf{Peak Learning Rate}                         & 1e-4 \\ 
\textbf{Adam $\epsilon$}                            & 1e-8 \\ 
\textbf{Adam ($\beta_1$, $\beta_2$)}                & (0.9, 0.999) \\ 
\textbf{Weight Decay}                               & 0.0  \\ 
\textbf{Learning Rate Decay Rate}                   & 0.75 \\
\textbf{Learning Rate Decay Step}                   & 20 \\
\bottomrule
\end{tabular}
\end{table*}

We report the detailed hyper-parameter settings used for training ExpC$^{\star}$ in Table \ref{tab:expc_details}.
The \#Layers, Hidden Dimension $d$, Dropout, and Batch Size are set to the default values provided in the official OGB demos without tuning.
The Expanding Hidden Dimension $d^{\prime}$ is set to 2 times of Hidden Dimension $d$ to reduce information loss in the aggregation step.
The learning rate is set to 1e-4, and a step learning rate scheduler is employed with a fixed decay rate of 0.75 every 20 epochs.

\section{Empirical Results}

\begin{table}[ht]
    \centering
    \begin{tabular}{c|c|c|c|c}
    \hline
      Model  &  Fold & Seed & Train MAE & Val MAE \\
      \hline
    \multirow{10}{*}{Graphormer} & 0 & 0 & 0.02054 & 0.0970 \\ 
     & 1 & 1 & 0.01809 & 0.0971 \\ 
     & 2 & 2 & 0.01897 & 0.0970 \\ 
     & 3 & 3 & 0.01833 & 0.0965 \\ 
     & 4 & 4 & 0.01817 & 0.0968 \\ 
     & 5 & 5 & 0.01964 & 0.0967 \\ 
     & 6 & 6 & 0.01941 & 0.0965 \\ 
     & 7 & 7 & 0.01832 & 0.0969 \\ 
     \cline{2-5}
     & All & 0 & 0.02146 & - \\ 
     & All & 1 & 0.02136 & - \\ 
     \hline
     \multirow{8}{*}{ExpC$^{\star}$}  & 0 & 0 & 0.06655 & 0.1009 \\ 
     & 1 & 1 & 0.06688 & 0.1008 \\ 
     & 2 & 2 & 0.07325 & 0.1012 \\ 
     & 3 & 3 & 0.07199 & 0.1017 \\ 
     & 4 & 4 & 0.07400 & 0.1004 \\ 
     & 5 & 5 & 0.05972 & 0.1018 \\ 
     & 6 & 6 & 0.07161 & 0.1003 \\
     & 7 & 7 & 0.06732 & 0.1028 \\
    \hline
    \end{tabular}
    \caption{Model Performance.}
    \label{tab:res}
\end{table}

Table \ref{tab:res} summarizes the performance of models used in the final submission. All models are trained by NVIDIA V100 GPUs on Microsoft Azure Cloud. 
8-fold cross validation strategy is adopted for training. As shown in Table \ref{tab:res}, model is trained using other $7$ folds and validated on the selected fold.
For the final submission, we leverage a naive weighted average to ensemble the logits of all models' output, where the coefficients are set manually by considering the performance of valid and train MAE, and the number of data used for training\footnote{We sum the logits of Graphormer using such weight: [0.05, 0.05, 0.05, 0.08, 0.05, 0.08, 0.08, 0.05, 0.05, 0.08], and ExpC$^{\star}$ using such weights (same order with Table 3): [0.05, 0.05, 0.05, 0.03, 0.05, 0.03, 0.05, 0.03], and then divide it by the summation of weights which is 0.96.}. Besides, we use PyTorch Geometric library\footnote{\url{https://pytorch-geometric.readthedocs.io/en/latest/}} to produce graph data from SMILES. However, we found that the different versions of PyTorch Geometric library may lead to a very slight difference of the produced data. For example, the sizes of {\tt geometric\_data\_processed.pt} produced by {\tt torch\_geometric=1.6.3} and {\tt torch\_geometric=1.7.0} are 8451179131 and 8451179119, respectively. The different graph data files would lead to a slight difference of the logits outputted by the same model. We conduct the inference for the final submission on multiple virtual machines. Therefore, a minor difference of logits (MAE <= 1e-4) is expected if the reproduction is conducted on a single machine.

\bibliographystyle{unsrtnat}
\bibliography{references}

\end{document}